\documentclass[10pt,twocolumn,letterpaper]{article}
\usepackage[table]{xcolor}
\usepackage{cvpr}
\usepackage{times}
\usepackage{epsfig}
\usepackage{graphicx}
\usepackage{amsmath}
\usepackage{amssymb}
\usepackage{fmtcount}

\usepackage[bookmarks=false]{hyperref}
\usepackage{amsmath}
\usepackage{amsthm}
\usepackage{amssymb}
\usepackage{epsfig}
\usepackage{url}
\usepackage{verbatim}
\usepackage{amsmath,amsfonts}
\usepackage[mathscr]{eucal}
\usepackage{amssymb,amsmath}
\usepackage{latexsym}
\usepackage{amsfonts}
\usepackage{amscd}
\usepackage{amsthm}
\usepackage{graphics}
\usepackage{algorithmic}
\usepackage{algorithm}

\newtheorem{theorem}{Theorem}[section]

\theoremstyle{definition}

\newcommand {\R}{\mathbb {R}}

\DeclareMathOperator{\argmin}{argmin}

\newcommand{\dist}{{\rm dist}}

\newcommand{\squishlist}{
 \begin{list}{$\bullet$}
  { \setlength{\itemsep}{0pt}
     \setlength{\parsep}{3pt}
     \setlength{\topsep}{3pt}
     \setlength{\partopsep}{0pt}
     \setlength{\leftmargin}{1.5em}
     \setlength{\labelwidth}{1em}
     \setlength{\labelsep}{0.5em} } }

\newcommand{\squishlisttwo}{
 \begin{list}{$\bullet$}
  { \setlength{\itemsep}{0pt}
     \setlength{\parsep}{0pt}
    \setlength{\topsep}{0pt}
    \setlength{\partopsep}{0pt}
    \setlength{\leftmargin}{2em}
    \setlength{\labelwidth}{1.5em}
    \setlength{\labelsep}{0.5em} } }

\newcommand{\squishend}{
  \end{list}  }

\DeclareMathOperator{\vol}{vol}

\newcommand{\reals}{\mathbb R}
\newcommand{\be}{\begin{equation}}
\newcommand{\ee}{\end{equation}}

\newcommand{\di}{{\,\mathrm{d}}}

\def\by{\mathbf{y}}
\def\bz{\mathbf{z}}

\def\sX{\mathrm{X}}
\def \di{\mathrm{d}}

\def\bx{\mathbf{x}}

\def\tb{\widetilde{\beta}}
\def\eps{\varepsilon}

\cvprfinalcopy 

\def\cvprPaperID{1879} 

\ifcvprfinal\fi
\begin{document}
\title{Randomized Hybrid Linear Modeling by Local Best-fit Flats\thanks{This work was supported by NSF grants
DMS-0612608, DMS-0811203 and DMS-0915064. Thanks to Peter Jones,
Mauro Maggioni and Amit Singer for some brief discussions that
motivated our exploration for a multiscale SVD-based HLM algorithm
and to the IMA for a stimulating multi-manifold modeling workshop.}}

\author{
Teng Zhang$^{\natural}$\hspace{0.5cm} Arthur Szlam$^{\flat}$
\hspace{0.5cm}Yi Wang$^{\natural}$ \hspace{0.5cm} Gilad
Lerman$^{\natural}$ \\
\begin{tabular}{cc}
   $\natural$\,School of Mathematics & $\flat$\,Courant Institute of
Mathematical Sciences \\
  University of Minnesota & New York University \\
{\tt\small\{zhang620,\,wangx857,\,lerman\}@umn.edu}& {\tt\small
aszlam@courant.nyu.edu} \\
\end{tabular}}


\maketitle

\begin{abstract}
The hybrid linear modeling problem is to identify a set of
$d$-dimensional affine sets in $\R^D$.  It arises, for example, in
object tracking and structure from motion.   The hybrid linear model
can be considered as the second simplest (behind linear) manifold
model of data.   In this paper we will present a very simple
geometric method for hybrid linear modeling based on selecting a set
of local best fit flats that minimize a global $\ell_1$ error
measure. The size of the local neighborhoods is determined
automatically by the Jones' $\beta_2$ numbers; it is proven under
certain geometric conditions that good local neighborhoods exist and
are found by our method.  We also demonstrate how to use this
algorithm for fast determination of the number of affine subspaces.
We give extensive experimental evidence demonstrating the state of
the art accuracy and speed of the algorithm on synthetic and real
hybrid linear data.
\end{abstract}

\noindent \textbf{Supp. webpage}:
http://www.math.umn.edu/$\sim$lerman/lbf/

\section{Introduction}

Many data sets can be modeled as unions of affine subspaces.  This
Hybrid Linear Modeling (HLM) finds diverse applications in many
areas, such as motion segmentation in computer vision, hybrid linear
representation of images, classification of face images, and
temporal segmentation of video sequences (see e.g.,
\cite{Vidal05,Ma07}).

Several algorithms have been suggested for solving this problem, for example
 the $K$-flats (KF) algorithm or any of its
variants~\cite{Kambhatla94fastnon-linear,Tipping99mixtures,Bradley00kplanes,Tseng00nearest,Ho03},
Subspace Separation~\cite{Costeira98,Kanatani01,Kanatani02},
Generalized Principal Component Analysis (GPCA)~\cite{Vidal05},
Local Subspace Affinity (LSA)~\cite{Yan06LSA}, Agglomerative Lossy
Compression and Spectral Curvature Clustering
(SCC)~\cite{spectral_applied}.  Some algorithms for modeling
data by a mixture of more general surfaces have been successfully
applied to HLM~\cite{LLMC,ssc09}.

In this paper, inspired by~\cite{Jones90,DS91,Lerman03}
and~\cite{MSVD09_1, MSVD09_2}, we will describe an extremely
straightforward geometric method for hybrid linear modeling that can
either be used in a stand alone manner or as an initialization of
any of the above methods. The basic idea is that for a data set
sampled from a hybrid linear model and a random point of it $\bx$:
the principal components of a neighborhood of appropriate size of
$\bx$ often give a good approximation to its nearest subspace. An
appropriate neighborhood size needs to be larger than the noise, so
that the affine cluster is recognized. However, not too large so
that the neighborhood intersects multiple clusters. Such
neighborhoods (in which a subspace is clearly distinguished) always
exist for points far enough from the intersection of subspaces
(i.e., most of points), as long as the following two assumptions are
satisfied: Samples are sufficiently dense along local regions of the
subspaces and data points sufficiently far from the intersection of
subspaces are mostly surrounded by neighbors of the same subspace
(this is true when the affine Grassmannian distance between
subspaces is sufficiently large and the noise level is sufficiently
small).

The contributions of this work are as follows: we make precise the
local fit heuristic, using the $\ell_2$ version of Jones' $\beta$
numbers~\cite{Jones90,DS91,Lerman03}, and state a theorem that tells
us under certain geometric conditions how to calculate the size of
the optimal local neighborhood.  Using this, we introduce a new
algorithm for affine clustering based on the above heuristic.  At
each of a randomly chosen subset of the data, we build a candidate
flat by calculating the principal components of a large neighborhood
which still lies in only one affine cluster.  The algorithm then
selects among the best fit flats of each of the neighborhoods to
build a global model using an $\ell_1$ error energy.  We show
experimentally that this algorithm obtains state of the art
accuracies on real and synthetic HLM problems while running
extremely fast
 (often on the order of ten times faster than most of the previously
mentioned methods).  Note that the two parts of the algorithm are
independent and can be used with other algorithms.  In particular,
we can use the local fit heuristic to initialize other HLM
algorithms.  We will give experimental evidence to show that the
$K$-flats algorithm~\cite{Ho03} is improved by such initialization.
We also show how to use this fast algorithm to quickly determine the
number of affine subspaces.

The rest of this paper is organized as follows. In
Section~\ref{sec:algorithm} we describe in greater depth the two
parts of the above algorithm, and state a theorem giving conditions
that guarantee that good neighborhoods can be found.
 Section~\ref{sec:exp} carefully tests the
algorithm on both artificial data of synthetic hybrid linear models
and real data of motion segmentation in video sequences. It also
demonstrates how to determine the number of clusters by applying the
fast algorithm of this paper together with the straightforward elbow
method. Section~\ref{sec:conclusions} concludes with a brief
discussion and mentions possibilities for future work.

\section{Randomized local best fit flats}
\label{sec:algorithm} The algorithm partitions a data set
$\mathrm{X}=\{\bx_1,\bx_2,\cdots,\bx_N\} \subseteq \mathbb{R}^{D}$
into $K$ clusters $\sX_1$, $\ldots$, $\sX_K$, with each cluster
approximated by a $d$-dimensional affine subspace, which we refer to
as $d$-flats or flats. We sketch it as follows, while suppressing
details that appear later in Algorithms~\ref{alg:nhbd}
and~\ref{alg:passes}.

\begin{algorithm}
\caption{HLM by randomized local best fit flats}
\label{alg:candidate}
\begin{algorithmic}
\REQUIRE $\sX=\{\bx_1,\bx_2,\cdots,\bx_n\} \subseteq
\mathbb{R}^{D}$: data, $d$: dimension of subspaces,  $C$: number of
candidate planes, $K$: number of output flats/clusters ($K<C$),
other parameters used by Algorithms~\ref{alg:nhbd}
and~\ref{alg:passes}
%
\ENSURE  A partition of $\sX$ into $K$ disjoint clusters
$\{\sX_i\}_{i=1}^K$, each approximated
by a single flat.\\
\textbf{Steps}:
\STATE
\squishlist 
    \item Pick $C$ random points in $\sX$
    \item For each of the $C$ points find appropriate local neighborhoods using Algorithm~\ref{alg:nhbd}
    \item Generate $C$ flats (by PCA) for the $C$ neighborhoods of the previous step
    \item Choose $K$ flats from the $C$ flats above using Algorithm~\ref{alg:passes}
    \item Partition $\sX$ by sending points to nearest $K$ flats
    above
\squishend 
\end{algorithmic}
\end{algorithm}

The proposed algorithm breaks into two main parts.  The first part
finds a set of candidate flats.  It takes as input the dimension of
the flats to be found and the number of candidates to search for. It
starts by randomly selecting one point for each candidate flat. The
algorithm chooses a scale (that is, a number of neighbors) around
each of the seed points. The best fit flats (in $L^2$ sense) for
each of the chosen neighborhoods are collected as candidates. The
method for choosing the best scale is described in
Section~\ref{sec:beta} and sketched in Algorithm~\ref{alg:nhbd}.

The second part of the algorithm searches for a good set of flats
from the candidates in a greedy fashion.  A number $K$ of desired
flats and a measure of goodness of a $K$ tuple of flats
$G=G_X(L_1,...L_K)$ is chosen; here, it will be the average $\ell_1$
distance of each point to its nearest flat.
After randomly initializing $K$ flats from the list of candidates,
$p$ passes are made through the data points.  One of the current
choices of flats is removed, and all the other candidates are tried
in its place.  If $G$ decreases, we replace the current flat with
the one which gives the lowest value for $G$.  We then move to the
next pass, picking a random flat, etc.

The simplest choice of $G$ is the sum of the squared distances of
each point in $X$ to its nearest flat.  In our experiments, we will
use an $\ell_1$ energy, i.e., summing the distance of every point to
its chosen flat.  We have experimentally found that this energy is
more robust to outliers than least squares error (see
also~\cite{MKF_workshop09} for a similar conclusion with a different
implementation of $\ell_1$ subspace minimization
and~\cite{lp_recovery09} for partial theoretical justification). One
can also imagine using spectral distances that measure the
smoothness of the clusters with respect to some kernel, or many
other global energy functionals of a partition.  The nice thing
about this method is that it allows for energy functionals which may
be hard to minimize; since we are only testing the energy of our
candidate configurations, as long as we can compute the energy of a
partition quickly, we can run the greedy descent.

\begin{algorithm}[t!]
\caption{Greedy $\ell_1$ candidate selection for HLM by randomized
local best fit flats} \label{alg:passes}
\begin{algorithmic}
\REQUIRE $\sX=\{\bx_1,\bx_2,\cdots,\bx_n\} \subseteq
\mathbb{R}^{D}$: data, $K$: number of flats, $L_1,...,L_C$:
candidate flats, and $p$: number of passes.
\ENSURE A set of $K$ ``active'' flats $\mathcal{L}\subset \{L_1,...,L_C\}$ .\\
%
\textbf{Steps}:
\STATE %
    Initialize $\mathcal{L}$ by randomly choosing $K$ ``active'' flats $L_{A_1},...,L_{A_K}$
    \FOR{$\text{pass}=1$ to $p$}
        \STATE
        Pick a random flat $L_{A_l} \subset \mathcal{L}$ ($1 \leq l \leq K$)
        \FOR{$j=1$ to $C-K$}
            \item $\bullet$ Pick one of the ``inactive'' flats $L_j$ and form the
            collection of flats $\tilde{\mathcal{L}}_j= L_j \bigcup  \mathcal{L}\setminus L_{A_l}$
            \item $\bullet$ Set $s_j=\sum_{i=1}^N \min_{L\in
            \tilde{\mathcal{L}}_j}{||x_i-P_Lx_i||}$
        \ENDFOR

        If $\min_j{s_j}<\sum_{i=1}^N \min_{L\in \{L_{A_1},...,L_{A_K}\}}{||x_i-P_Lx_i||}$,
        set $L_{A_l}:= L_{\argmin{s_j}}$
    \ENDFOR
%
\end{algorithmic}
\end{algorithm}

\subsection{Choosing the optimal neighborhood}
\label{sec:beta}

\begin{algorithm}[!t]
\caption{Neighborhood size selection for HLM by randomized local
best fit flats} \label{alg:nhbd}
\begin{algorithmic}
\REQUIRE $\sX=\{\bx_1,\bx_2,\cdots,\bx_n\} \subseteq
\mathbb{R}^{D}$: data, $\bx$: a point in $\sX$, $S$: start size,
$T$: step size, $\ell, m$ (optional): mean shifts parameters.
\ENSURE $\mathcal{N}(\bx)$: a neighborhood of $\bx$.\\
\textbf{Steps}:
\STATE
%
     $\bullet$ (Optional) Update the point $\bx$ as the center of its $\ell$-nearest neighborhood in $\sX$, while
     repeating $m$      times\\
     $\bullet$ $k=-1$
    \REPEAT \STATE
     $\bullet$ k:=k+1\\
     $\bullet$ Set $\mathcal{N}_k$ to be the $S+kT$ nearest points in $\sX$ to
     $\bx$\\
     $\bullet$ Set $\tilde{L}_k$ to be the best fit flat to $\mathcal{N}_k$\\
     $\bullet$ Compute $\beta_2(k):=\beta_2(\mathcal{N}_k)$ according to~\eqref{eq:def_beta}
     \UNTIL $k>1$ and
    $\beta_2(k-1)<\min\{\beta_2(k-2),\beta_2(k)\}$\\
    $\bullet$ Output $\mathcal{N}(\bx) :=\mathcal{N}_{k-1}$
%
\end{algorithmic}
\end{algorithm}

Choosing the correct neighborhood is crucial for the success of the
method.  If the neighborhood is too small, even if the point is in a
good affine cluster, then a small amount of noise in the data will
result in a flat which does not match most of the points in the
affine cluster. If the neighborhood is too large, it will contain
points from more than one affine cluster, and the resulting best fit
flat will again not match any of the actual data points.  While it
is possible to take a guess at the correct scale as a parameter, we
have found that it is possible to choose the correct scale
reasonably well automatically.

What we will do is start at the smallest scale (say $d+1$) and look
at larger and larger neighborhoods of a given point $\bx_0$. At the
smallest scale, any noise causes the local neighborhood to look $D$
dimensional.  As we add points to the neighborhood, it becomes
better and better approximated in an average sense by its best fit
flat, until points belonging to other flats enter the neighborhood.
We thus take the neighborhood which is the first local minimum of
the scaled least squares error for $d$-flat approximation. In
practice, for a neighborhood $\mathcal{N}$ of $\bx_0$ the scaled
least squares error for $d$-flat approximation,
$\beta_2(\mathcal{N})$, is computed by the formula:
\be \label{eq:def_beta}
\beta_2(\mathcal{N})=\left(\min_{\text{$d$-flats
$L$}}\frac{\sum_{\by\in \mathcal{N}} ||\by-P_{L}\by||^2}
{|\mathcal{N}|(\max_{\bx\in
\mathcal{N}}||\bx-\bx_0||)^2}\right)^\frac{1}{2}, \ee
where $P_{L}$ denotes the projection onto the flat $L$. This notion
of scaled error introduced and utilized
in~\cite{Jones90,DS91,Lerman03}, and considered recently in
\cite{MSVD09_1, MSVD09_2} for dimension estimation. The procedure we
have just described is summarized in Algorithm~\ref{alg:nhbd}.

 The following theorem tries to justify
our strategy of fitting the correct scale around each point. We work
with a ``geometric'' set of assumptions in the continuous setting,
where our data set will be presumed to be a collection of tubes
around flats.  This corresponds roughly to a probabilistic setting
of sampling according to mixtures of uniform distributions around
subsets of $d$-flats.  For convenience we assume infinite tubes but
restrict to local scales.

 The analog of the discrete $\beta_2$
introduced earlier when having an underlying continuous set $\Omega$
(here it is the union of tubes) in a ball of center $\bx$ and radius
$r$ is defined as follows:
\be \nonumber \beta_2^2(\bx,r) = \min_{L} \int\limits_{\Omega \cap
B(\bx,r)} \left(\frac{\dist(\bx,L)}{2 r}\right)^2 \frac{\di
\bx}{\vol(\Omega \cap B(\bx,r))}\, \ee
where the minimum is over all $d$-flats $L$ (see
also~\cite{Lerman03}).

\begin{theorem}
Let $K \geq 2$, $d<D$ , $L_i$, $i=1, \ldots, K$, be $K$ $d$-flats in
$\reals^D$, and $\Omega_i:=T(L_i,w_i)$  be $K$ tubes in $\reals^D$
around these flats of comparable widths $\{w_i\}_{i=1}^K$.

For fixed $1 \leq i^* \leq K$ and fixed $\bx\in L_{i^*}$, let
%
\be\by= \by(\bx) =
\argmin\limits_{\by\in\Omega\setminus\Omega_{i^*}} \dist(\by,\bx)\ee
and
\be r_0:=\dist(\by,\bx).\ee
Assume that $r_0
> w_{i^*}$. Then
the function $\beta_2(\bx, r)$ is constant for $r$ in $[0,w_{i^*}]$,
comparable to a function which is decreasing for a sufficiently
large subinterval of $[w_{i^*}, r_0]$, and satisfies the inequality
\begin{equation}
\label{eq:beta_local_min} \beta_2((1+\eps)\cdot r_0) \gtrapprox
\beta_2(r_0)
\end{equation}
%
%
for sufficiently small $\eps$, i.e., it has an ``approximate'' local
minimum in the interval $[r_0, (1+\eps) \cdot r_0]$. If $d \leq 4$,
then $\eps \approx {w_{i^*}}/{r_0}$, and if $d > 4$ then $\eps
\approx \left( {w_{i^*}}/{r_0} \right)^{{4}/{d}}$. As $w_{i^*}/r_0$
approaches zero, all comparability constants mentioned above
approach one.
\end{theorem}

We remark that by imposing an upper bound on the widths of the tubes
in the theorem above and a lower bound on the dihedral angles
between the flats, then the local condition $r_0
> w_{i^*}$ (required by the theorem) is satisfied at any point $\bx$ which has distance larger than
order of $\max_{1 \leq i \leq K} w_i$ from the intersection of all
flats.

\subsection{Some technical notes about the proposed algorithm}
\label{sec:eng}
Note that the first minimum in the Theorem excludes the left
endpoint.  In our experiments, we noticed that on data without too
much noise, it is useful to allow the first scale to count as a
local minimum.  In the experiments below, we will show the results
of the algorithm with both notions of "first" local minimum.

The second technical detail concerns the choice of the random points
used for candidate generation.  We use the mean shift technique:
given a point $\bx$, update $\bx$ as the center of its neighborhood
several times. The method
 shifts the point to a denser region, resulting in a more accurate estimation
 of the flats.   In the experiments below, we will show the results with and
without mean shift biased seed selection.

\section{Experimental results}
\label{sec:exp}

\addtocounter{footnote}{1}
\begin{table*}[htbp]

\centering \caption{\small {Mean percentage of misclassified points
in simulation for linear-subspace cases or affine-subspace case.
   The proposed algorithm as in
Section~\ref{sec:eng} is in the row labeled LBFMS, and the
``vanilla'' version is in the row labeled by LBF}}\label{tab:error}
{\scriptsize
\begin{tabular}{|r||r|r||r|r||r|r||r|r||r|r||r|r||}
 \hline
   \cline{2-11}
    &\multicolumn{2}{c||}{}
    &\multicolumn{2}{c||}{}
    &\multicolumn{2}{c||}{}
    &\multicolumn{2}{c||}{}
    &\multicolumn{2}{c||}{$(4,5,6)$}\\
    \raisebox{1.5ex}[0pt]{\normalsize{Linear}}
    &\multicolumn{2} {c||}{\raisebox{1.5ex}[0pt]{$2^2 \in\mathbb{R}^4$}}
    &\multicolumn{2}{c||}{\raisebox{1.5ex}[0pt]{$4^2 \in\mathbb{R}^6$}}
    &\multicolumn{2}{c||}{\raisebox{1.5ex}[0pt]{$2^4 \in\mathbb{R}^4$}}
    &\multicolumn{2}{c||}{\raisebox{1.5ex}[0pt]{$10^2 \in\mathbb{R}^{15}$}}
    &\multicolumn{2}{c||}{$\in\mathbb{R}^{10}$}\\
    \cline{2-11}

    \cline{1-11}
      Outl.  \%       & 5 & 30 & 5 & 30 & 5 & 30 & 5 & 30 & 5 & 30\\
  \hline\hline
LSCC&  3.0&6.9&2.3&2.6&7.7&22.4&0.5&3.8&1.8&28.2\\
 \rowcolor[gray]{.8}LSA&18.7&19.6&10.9&12.7&44.3&21.0&7.6&9.9&6.1&6.6\\
KF&3.0&15.8&2.5&18.4&9.4&34.3&0.8&33.8&0.8&30.6\\
 \rowcolor[gray]{.8}MoPPCA&3.1&14.2&2.5&17.7&8.4&34.2&0.9&38.8&1.4&34.7\\
GPCA&19.7&30.9&11.7&35.9&29.2&43.9&10.2&42.6&10.1&45.4\\

 \rowcolor[gray]{.8}LBF&2.7&3.0&2.7&2.6&7.0&11.1&1.5&2.1&1.4&1.9\\
LBFMS&3.1&3.0&2.7&2.8&7.0&11.3&4.3&5.5&2.1&1.9\\
 \rowcolor[gray]{.8} RANSAC$^{\decimal{footnote}}$&3.3&2.6&2.3&2.2&8.6&9.8&0.9&6.7&1.8&1.4\\

  \hline
\end{tabular}

\begin{tabular}{|r||r|r||r|r||r|r||r|r||r|r||r|r||}
 \hline
  \cline{2-11}
    &\multicolumn{2}{c||}{}
    &\multicolumn{2}{c||}{}
    &\multicolumn{2}{c||}{}
    &\multicolumn{2}{c||}{}
    &\multicolumn{2}{c||}{$(4,5,6)$}\\
    \raisebox{1.5ex}[0pt]{\normalsize{Affine}}
    &\multicolumn{2} {c||}{\raisebox{1.5ex}[0pt]{$2^2 \in\mathbb{R}^4$}}
    &\multicolumn{2}{c||}{\raisebox{1.5ex}[0pt]{$4^2 \in\mathbb{R}^6$}}
    &\multicolumn{2}{c||}{\raisebox{1.5ex}[0pt]{$2^4 \in\mathbb{R}^4$}}
    &\multicolumn{2}{c||}{\raisebox{1.5ex}[0pt]{$10^2 \in\mathbb{R}^{15}$}}
    &\multicolumn{2}{c||}{$\in\mathbb{R}^{10}$}\\
    \cline{2-11}

    \cline{1-11}
          Outl.  \%       & 5 & 30 & 5 & 30 & 5 & 30 & 5 & 30 & 5 & 30\\
  \hline\hline
SCC& 0.0&0.6&0.0&0.0&0.2&0.5&0.0&0.7&0.0&5.8\\
 \rowcolor[gray]{.8} LSA&11.8&11.0&5.3&4.7&45.0&41.7&0.0&0.0&1.0&1.1\\
KF&7.3&15.1&9.9&26.0&19.7&37.1&11.1&24.9&7.3&23.5\\
 \rowcolor[gray]{.8} MoPPCA&25.6&23.7&27.8&38.3&45.5&39.8&37.1&45.2&42.9&46.8\\
GPCA&13.8&14.4&22.6&22.1&33.6&32.4&36.0&29.6&26.7&29.1\\
 \rowcolor[gray]{.8} LBF&0.2&2.1&0.1&1.8&0.5&3.7&0.0&0.5&0.0&0.0\\

LBFMS&0.4&2.0&0.1&2.6&0.7&6.0&0.0&0.3&0.0&0.0\\

 \rowcolor[gray]{.8} RANSAC$^{\decimal{footnote}}$
 &13.2&12.2&11.5&11.2&31.5&28.4&2.6&9.2&1.1&2.2\\

  \hline
\end{tabular}
}
\end{table*}

\begin{table*}[htbp]

\centering \caption{\small {Mean running time for linear-subspaces
cases and affine-subspaces cases.   The proposed algorithm as in
Section~\ref{sec:eng} is in the row labeled LBFMS, and the
``vanilla'' version is in the row labeled by LBF.}}\label{tab:time}
{\scriptsize
\begin{tabular}{|r||r|r||r|r||r|r||r|r||r|r||}
 \hline
  \cline{2-11}
    &\multicolumn{2}{c||}{}
    &\multicolumn{2}{c||}{}
    &\multicolumn{2}{c||}{}
    &\multicolumn{2}{c||}{}
    &\multicolumn{2}{c||}{$(4,5,6)$}\\
    \raisebox{1.5ex}[0pt]{\normalsize{Linear}}
    &\multicolumn{2} {c||}{\raisebox{1.5ex}[0pt]{$2^2 \in\mathbb{R}^4$}}
    &\multicolumn{2}{c||}{\raisebox{1.5ex}[0pt]{$4^2 \in\mathbb{R}^6$}}
    &\multicolumn{2}{c||}{\raisebox{1.5ex}[0pt]{$2^4 \in\mathbb{R}^4$}}
    &\multicolumn{2}{c||}{\raisebox{1.5ex}[0pt]{$10^2 \in\mathbb{R}^{15}$}}
    &\multicolumn{2}{c||}{$\in\mathbb{R}^{10}$}\\
    \cline{2-11}

    \cline{1-11}
      Outl.  \%       & 5 & 30 & 5 & 30 & 5 & 30 & 5 & 30 & 5 & 30\\
  \hline\hline
LSCC&  0.7&0.8&16.0&1.8&2.1&2.0&13.3&5.7&5.1&8.4\\
 \rowcolor[gray]{.8} LSA&8.8&16.0&11.1&20.8&28.3&54.4&31.3&31.5&38.2&54.4\\
KF&0.5&0.6&0.5&0.8&1.4&1.8&1.9&1.0&1.1&2.8\\
 \rowcolor[gray]{.8} MoPPCA&0.2&0.5&0.3&0.7&1.2&2.0&1.7&1.1&1.0&3.3\\
GPCA&3.5&7.6&9.8&19.0&20.9&29.7&30.3&31.6&39.1&57.8\\
 \rowcolor[gray]{.8} LBF&0.3&0.3&0.3&0.3&0.9&1.1&0.6&0.6&0.6&0.8\\
LBFMS&0.3&0.3&0.3&0.3&1.1&1.4&0.4&0.5&0.7&0.9\\
 \rowcolor[gray]{.8} RANSAC$^{\decimal{footnote}}$&0.01&0.01&0.02&0.06&0.03&0.06&3.5&3.8&0.9&3.4\\

  \hline
\end{tabular}

\begin{tabular}{|r||r|r||r|r||r|r||r|r||r|r||r|r||}
 \hline
  \cline{2-11}
    &\multicolumn{2}{c||}{}
    &\multicolumn{2}{c||}{}
    &\multicolumn{2}{c||}{}
    &\multicolumn{2}{c||}{}
    &\multicolumn{2}{c||}{$(4,5,6)$}\\
    \raisebox{1.5ex}[0pt]{\normalsize{Affine}}
    &\multicolumn{2} {c||}{\raisebox{1.5ex}[0pt]{$2^2 \in\mathbb{R}^4$}}
    &\multicolumn{2}{c||}{\raisebox{1.5ex}[0pt]{$4^2 \in\mathbb{R}^6$}}
    &\multicolumn{2}{c||}{\raisebox{1.5ex}[0pt]{$2^4 \in\mathbb{R}^4$}}
    &\multicolumn{2}{c||}{\raisebox{1.5ex}[0pt]{$10^2 \in\mathbb{R}^{15}$}}
    &\multicolumn{2}{c||}{$\in\mathbb{R}^{10}$}\\
    \cline{2-11}

    \cline{1-11}
      Outl.  \%       & 5 & 30 & 5 & 30 & 5 & 30 & 5 & 30 & 5 & 30\\
  \hline\hline
SCC&0.9&1.0&1.7&2.0&5.1&2.5&6.1&13.7&5.6&6.0\\
 \rowcolor[gray]{.8} LSA&8.7&16.1&11.1&20.8&28.6&54.0&21.1&32.2&38.3&54.0\\
KF&0.5&0.6&0.6&0.7&2.4&1.4&0.6&1.7&1&1.4\\
  \rowcolor[gray]{.8} MoPPCA&0.5&0.5&0.7&0.6&2.9&1.4&1.3&1.9&1.9&2.0\\
GPCA&2.4&6.9&5.1&9.8&11.2&26.1&20.2&31.9&38.4&49.9\\
 \rowcolor[gray]{.8} LBF& 0.3&0.3&0.3&0.3&1.1&1.3&0.5&0.6&0.7&0.9\\

LBFMS&0.3&0.3&0.3&0.3&1.1&1.5&0.4&0.5&0.7&0.9\\
 \rowcolor[gray]{.8} RANSAC$^{\decimal{footnote}}$&0.02&0.1&0.2&0.6&0.2&0.3&3.2&3.7&2.0&3.5\\

  \hline
\end{tabular}
}
\end{table*}

In this section, we conduct experiments on artificial and real data
sets to verify the effectiveness of the proposed algorithm in
comparison to other hybrid linear modeling algorithms.

We measure the accuracy of those algorithms by the rate of
misclassified points with outliers excluded, that is
\begin{equation}
\text{error}\%= \frac{\text{\# of misclassified inliers}}{\text{\#
of total inliers}} \times 100\%\,.
\end{equation}

In all the experiments below, the number $C$ in
Algorithm~\ref{alg:candidate} is 70 times the number of subspaces,
the number $p$ in Algorithm~\ref{alg:passes} is 3 times the number
of subspaces, and the number $T$ in Algorithm~\ref{alg:nhbd} is 2.
We run experiments with and without mean shifts; the experiments
using mean shifts use 10-nearest neighbors and 5 shifts.  According to our experience the LBF algorithm is very robust to changes in parameters, but unsurprisingly, there is a general trade off between
accuracy (higher $C$, higher $p$, smaller $T$), and run time (lower
$C$, lower $p$, larger $T$).  We have chosen these parameters for a
balance between run time and accuracy.

\subsection{Simulated data}
\label{sec:sim}

\footnotetext[\value{footnote}]{The RANSAC code we use (and most standard
versions of RANSAC) depend on a user supplied inlier threshold.  The first
part of our algorithm can in some sense be considered to be the automatic detection
of this inlier threshold; and if this is provided by the user, the initialization
we have described is no longer useful, as we would simply pick the largest neighborhood so that
the distance from any point to its projection is smaller than the user supplied bound.
The experiments in the table use the oracle choice of inlier bound (given by the true
noise variance), and so here
RANSAC has an advantage over the other algorithms listed.}

We compare our algorithm with the following algorithms: Mixtures of
PPCA (MoPPCA)~\cite{Tipping99mixtures}, $K$-flats (KF)~\cite{Ho03},
Local Subspace Analysis (LSA)~\cite{Yan06LSA}, Spectral Curvature
Clustering (SCC)~\cite{spectral_applied}, Random Sample Consensus
(RANSAC)~\cite{Fischler81RANSAC} and GPCA with voting
(GPCA)~\cite{Ma07}. We use the Matlab codes of the GPCA, MoPPCA and
KF algorithm from http://percep
tion.csl.uiuc.edu/gpca, the SCC
algorithm from http://www
.math.umn.edu/$\sim$lerman/scc and the
LSA, RANSAC algorithms from http://www.vision.jhu.edu/db.

The MoPPCA algorithm is always initialized with a random guess of
the membership of the data points. The LSCC algorithm is initialized
by randomly picking $100\times K$ $(d+1)$-tuples (following
~\cite{spectral_applied}), and KF
 are
initialized with random guess. Since algorithms like KF tend to
converge to local minimum, we use 10 restarts for MoPPCA, 30
restarts for KF, and recorded the misclassification rate of the one
with the smallest $\ell_2$ error for MoPPCA as well as KF. The
number of restarts was restricted by the running time and accuracy.
RANSAC uses the oracle inlier bound given by the model's noise
variance.

\begin{table*}[htbp!]

\centering \caption{\label{tab:comp1}  \small The mean and median
percentage of misclassified points for two-motions and three-motions
in Hopkins 155 database.  The proposed algorithm as in
Section~\ref{sec:eng} is in the row labeled LBFMS, and the
``vanilla'' version is in the row labeled by LBF} \vspace{.1in}
{\scriptsize
\begin{tabular}{|l||r|r||r|r||r|r||r|r|}
  \hline

    &\multicolumn{2}{c||}{Checker}
    &\multicolumn{2}{c||}{Traffic}
    &\multicolumn{2}{c||}{Articulated}
    &\multicolumn{2}{c|}{All}\\
    \cline{2-9}
    \raisebox{1.5ex}[0pt]{\normalsize{2-motion}} &Mean &Median &Mean &Median &Mean &Median &Mean &Median\\
  \hline\hline

  GPCA     & 6.09&1.03&1.41&0.00&2.88&0.00&4.59&0.38   \\



 \rowcolor[gray]{.8} LLMC 5&4.37&0.00&0.84&0.00&6.16&1.37&3.62&0.00\\
LSA 4$K$&  2.57&0.27&5.43&1.48&4.10&1.22&3.45&0.59\\
 \rowcolor[gray]{.8} LBF(4$K$,3)&3.31&0.00&3.29&0.00&4.31&0.12&3.40&0.00\\
LBFMS(4$K$,3)&3.05&0.00&0.78&0.00&1.73&0.03&2.34&0.00\\
 \rowcolor[gray]{.8}MSL&4.46&0.00&2.23&0.00&7.23&0.00&4.14&0.00\\
RANSAC&6.52&1.75&2.55&0.21&7.25&2.64&5.56&1.18\\

 \rowcolor[gray]{.8} SCC(4$K$,4)&1.30&0.04&1.07&0.44&3.68&0.44&1.46&0.16\\
SSC-N& 1.12&0.00&0.02&0.00&0.62&0.00&0.82&0.00\\



  \hline

\end{tabular}

\begin{tabular}{|l||r|r||r|r||r|r||r|r|}
  \hline

    &\multicolumn{2}{c||}{Checker}
    &\multicolumn{2}{c||}{Traffic}
    &\multicolumn{2}{c||}{Articulated}
    &\multicolumn{2}{c|}{All}\\
    \cline{2-9}
    \raisebox{1.5ex}[0pt]{\normalsize{3-motion}}  &Mean &Median &Mean &Median &Mean &Median &Mean &Median\\
  \hline\hline
GPCA&  31.95&32.93&19.83&19.55&16.85&28.66&28.66&28.26\\


 \rowcolor[gray]{.8} LLMC 4$K$&12.01&9.22&7.79&5.47&9.38&9.38&11.02&6.81\\

LLMC 5&10.70&9.21&2.91&0.00&5.60&5.60&8.85&3.19\\
 \rowcolor[gray]{.8} LSA 4$K$&5.80&1.77&25.07&23.79&7.25&7.25&9.73&2.33\\
LSA 5&30.37&31.98&27.02&34.01&23.11&23.11&29.28&31.63\\
 \rowcolor[gray]{.8} LBF(4$K$,3)&8.42&1.29&14.80&9.21&20.45&20.45&10.38&1.63\\
LBFMS(4$K$,3)&6.87&1.47&1.40&0.00&24.10&24.10&6.76&0.89\\
 \rowcolor[gray]{.8} MSL&10.38&4.61&1.80&0.00&2.71&2.71&8.23&1.76\\
RANSAC&25.78&26.01&12.83&11.45&21.38&21.38&22.94&22.03\\

 \rowcolor[gray]{.8}SCC(4$K$,4)&5.68&2.96&2.35&2.07&10.94&10.94&5.31&2.40\\
SSC-N&2.97&0.27&0.58&0.00&1.42&0.00&2.45&0.20\\


  \hline
\end{tabular}
}
\end{table*}

The simulated data represents various instances of $K$ linear
subspaces in $\mathbb{R}^D$. If their dimensions are fixed and equal
$d$, we follow~\cite{spectral_applied} and refer to the setting as
$d^K\in \mathbb{R}^D$. If they are mixed, then we follow~\cite{Ma07}
and refer to the setting as $(d_1, \ldots, d_K) \in \mathbb{R}^D$.
Fixing $K$ and $d$ (or $d_1, \ldots, d_K$), we randomly generate 100
different instances of corresponding hybrid linear models according
to the code in http://perception.csl.uiuc.edu/gpca. More precisely,
for each of the 100 experiments, $K$ linear subspaces of the
corresponding dimensions in $\reals^D$ are randomly generated.
Within each subspace, the underlying sampling distribution is the sum of a uniform distribution in a $d$-dimensional ball of
radius $1$ of that subspace (centered at the origin for the case of linear subspaces) and a $D$-dimensional multivariate normal
distribution with mean $\mathbf{0}$ and covariance matrix $0.05^2\cdot \mathbf{I}_{D\times D}$. Then,
for each subspace 250 samples are generated according to the
distribution just described. Next, the data is further corrupted
with 5\% or 30\% uniformly distributed outliers in a cube of
sidelength determined by the maximal distance of the former 250
samples to the origin (using the same code).

Since most algorithms (including ours) do not support mixed
dimensions natively, we assume each subspace has the maximum
dimension in the experiment.

The mean (over 100 instances) misclassification rate of the various
algorithms is recorded in Table~\ref{tab:error}. The mean running
time is shown in Table~\ref{tab:time}.  In each of the Tables, our
algorithm is labeled LBF (Local Best-fit Flats); our algorithm with
mean shifts and using the modified choice of good neighborhood
described in section \ref{sec:eng} is labeled LBFMS.

\subsection{Motion segmentation data}

We test the proposed algorithm on the Hopkins 155 database of motion
segmentation, which is available at
http://www.vision.jhu.edu/data/hopkins155. This data contains 155
video sequences along with the coordinates of certain features
extracted and tracked for each sequence in all its frames. The main
task is to cluster the feature vectors (across all frames) according
to the different moving objects and background in each video.

More formally, for a given video sequence, we denote the number of
frames by $F$. In each sequence, we have either one or two
independently moving objects, and the background can also move due
to the motion of the camera. We let $K$ be the number of moving
objects plus the background, so that $K$ is 2 or 3 (and distinguish
accordingly between two-motions and three-motions). For each
sequence, there are also $N$ feature points
$\by_1,\by_2,\cdots,\by_N \in \mathbb{R}^3$ that are detected on the
objects and the background. Let $\bz_{ij}\in \mathbb{R}^2$ be the
coordinates of the feature point $\by_j$ in the $i^{th}$ image frame
for every $1\leq i\leq F$ and $1\leq j \leq N$. Then
$\bz_j=[\bz_{1j},\bz_{2j},\cdots,\bz_{Fj}] \in \mathbb{R}^{2F}$ is
the trajectory of the $j^{th}$ feature point across the $F$ frames.
The actual task of motion segmentation is to separate these
trajectory vectors $\bz_1,\bz_2,\cdots,\bz_N$ into $K$ clusters
representing the $K$ underlying motions.

\begin{figure}
\begin{center}
\includegraphics[width=.4\textwidth,height=.3\textwidth]{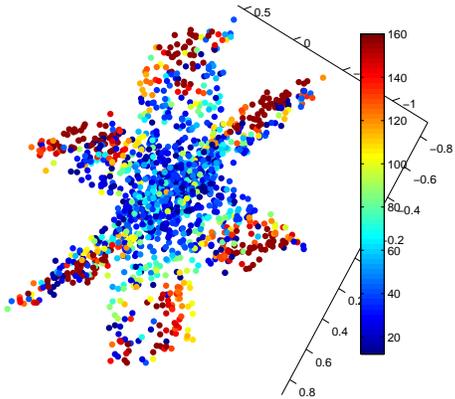}
\caption{\it Data set \#3 from Section~\ref{sec:ikf}.  The color
value represents the number of neighbors chosen at that point. Note
that the algorithm chooses smaller neighborhoods for points closer
to the intersection of the planes. \label{fig:nonei}}
\end{center}
\end{figure}

\begin{figure*}[htbp]
\begin{center}
\begin{minipage}{0.31\textwidth} \centering
\includegraphics[width=1\textwidth,height=.8\textwidth]{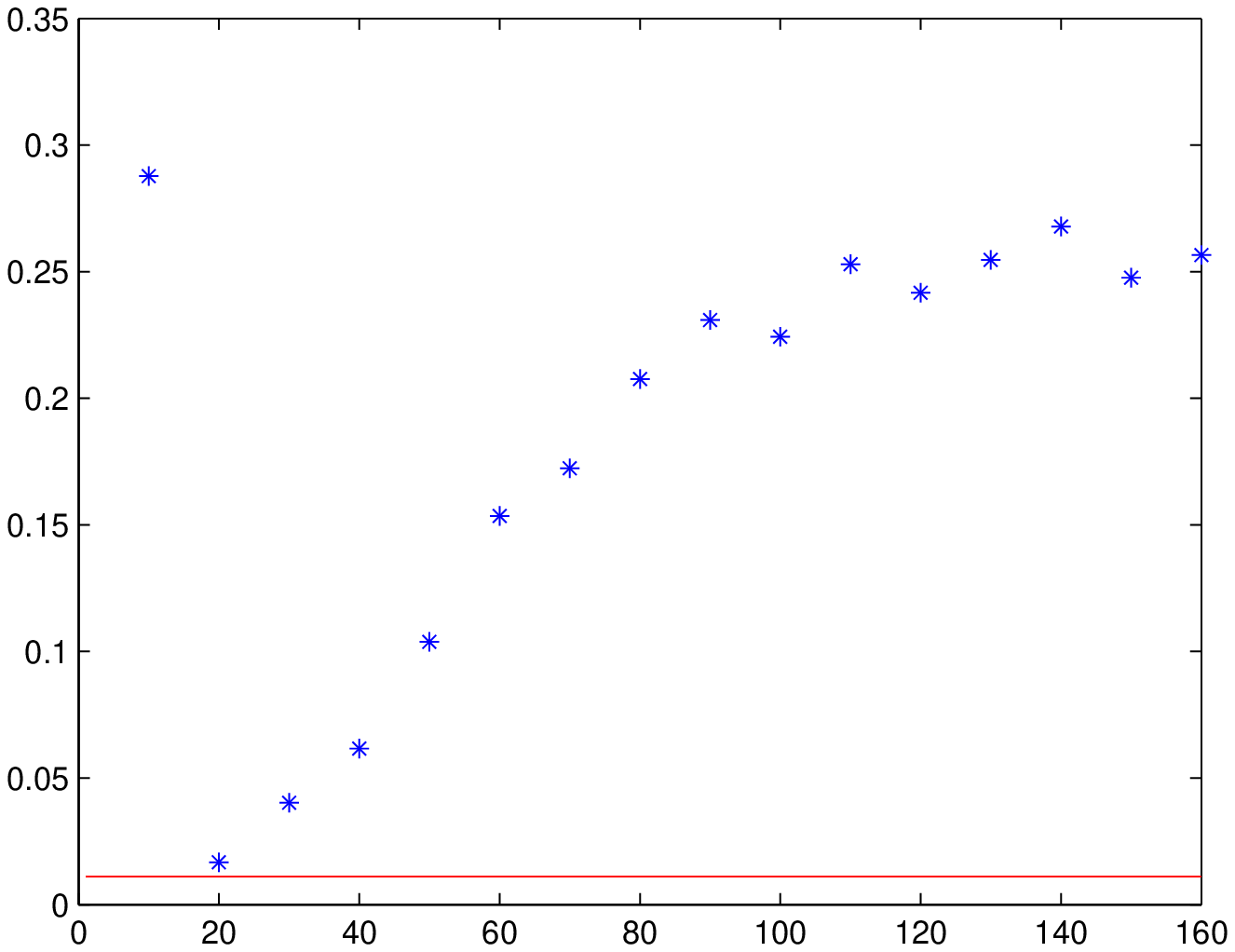}
\end{minipage}
\begin{minipage}{0.31\textwidth} \centering
\includegraphics[width=1\textwidth,height=.8\textwidth]{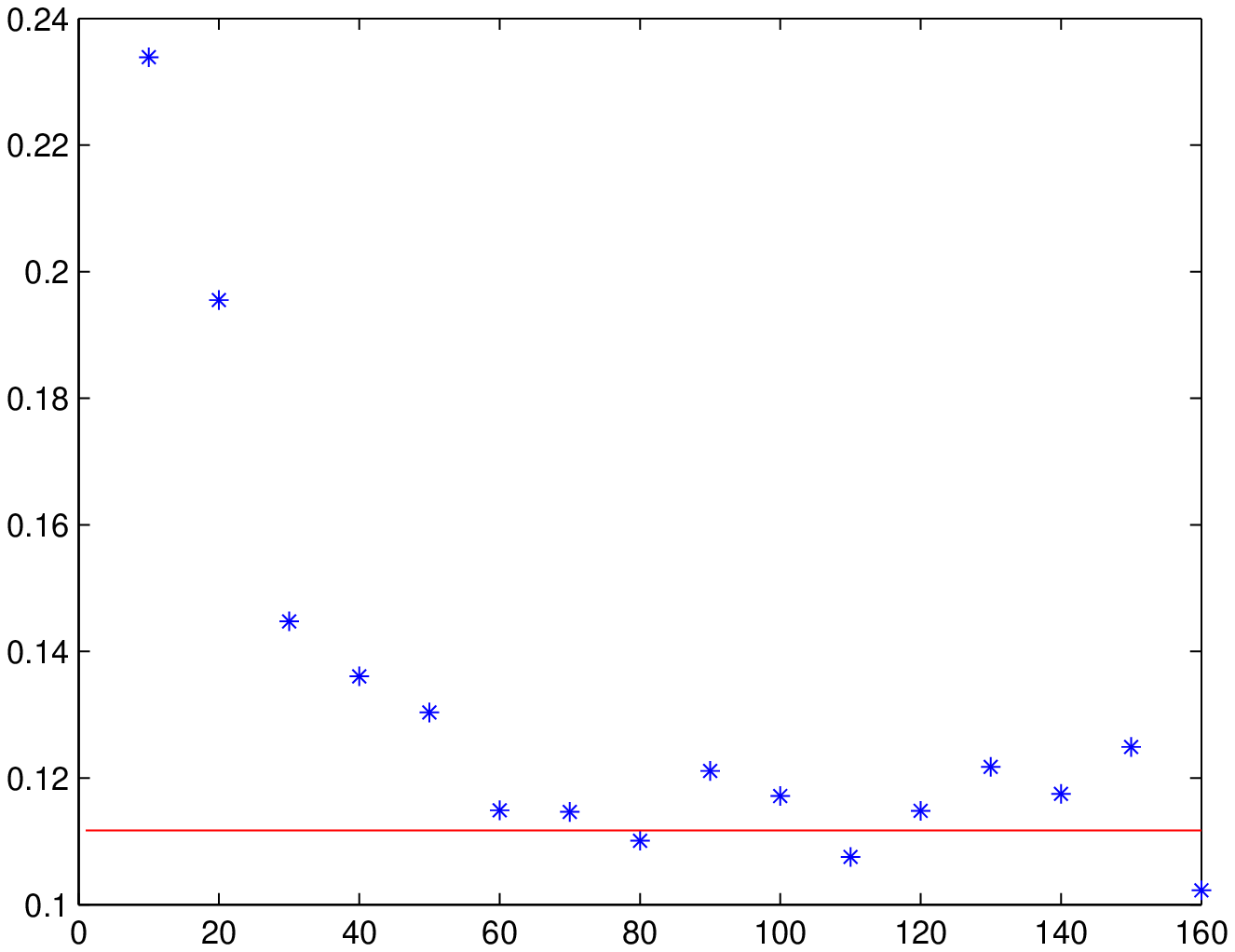}
\end{minipage}
\begin{minipage}{0.31\textwidth} \centering
\includegraphics[width=1\textwidth,height=.8\textwidth]{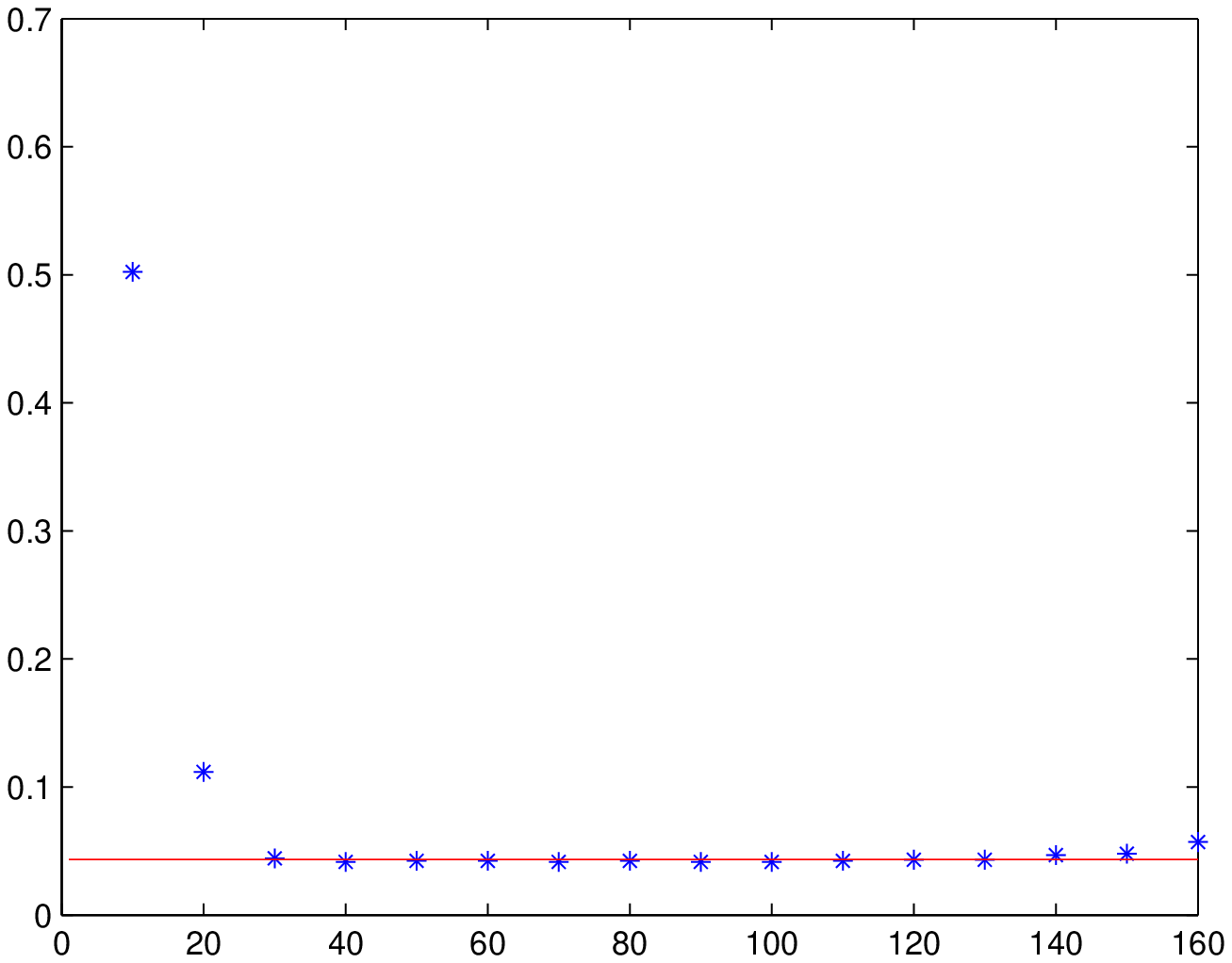}
\end{minipage}
\caption{\it Using our neighborhood choice to improve initialization
of $k$-flats: the vertical axis is accuracy, and the horizontal axis
is fixed neighborhood size in geometric farthest insertion for
initialization of $K$ flats.  The red line is the result of using
adapted neighborhoods. The data sets are \#1,\#2, and \#3 as
described in Section~\ref{sec:ikf}.  Random initialization leads to
errors of .4 or greater for all three data sets.
\label{fig:kmeansit} }
\end{center}
\end{figure*}

It has been shown~\cite{Costeira98} that under affine camera models
and with some mild conditions, the trajectory vectors corresponding
to different moving objects and the background across the $F$ image
frames live in distinct affine subspaces of dimension at most three
in $\mathbb{R}^{2F}$. Following this theory, we implement our
algorithm with $d=3$, and use affine flats.


\setlength{\abovecaptionskip}{4pt}   
\setlength{\belowcaptionskip}{4pt}
\begin{table*}[htbp]
\centering \caption{\small {The percentage of incorrectness ($e\%$)
and the average computation time $t$ of the three methods SOD (LBF),
ALC and GPCA.}}\label{tab:SOD} {\scriptsize
\begin{tabular}{|c|c|@{}c@{}|@{}c@{}|@{}c@{}|@{}c@{}|@{}c@{}|@{}c@{}|@{}c@{}|@{}c@{}|@{}c@{}|@{}c@{}|@{}c@{}|}
\hline

&&\multicolumn{7}{c|}{no minimum angle}&\multicolumn{4}{c|}{minimum
angle
$=\pi/8$} \\
\hline

&&$1^6\in\mathbb{R}^5$ &$2^4\in\mathbb{R}^5$ &$3^3\in\mathbb{R}^5$
&$10^2\in\mathbb{R}^{15}$ &$1^6\in\mathbb{R}^3$
&$2^4\in\mathbb{R}^3$ &$3^3\in\mathbb{R}^4$ &$1^6\in\mathbb{R}^3$
&$2^4\in\mathbb{R}^3$ &$3^3\in\mathbb{R}^4$
&$10^2\in\mathbb{R}^{15}$ \\

\hline

&e\%&17&3&2&0&55&29&19&3&5&5&0\\

\raisebox{1.5ex}[0pt]{SOD (LBF)}&t&3.51&4.07&3.37&7.31&3.13&3.77&3.85&3.09&3.45&3.32&6.78\\
\hline

ALC&e\%&1&0&0&16&34&31&1&0&10&1&13\\

$\epsilon=0.05$&t&23.74&43.44&59.14&1370.92&20.49&37.49&53.59&20.22&37.41&54.11&1354.11\\
\hline

&e\%&88&100&100&100&27&100&100&13&100&100&100\\

\raisebox{1.5ex}[0pt]{GPCA}&t&0.03&0.09&0.12&1.30&0.06&0.09&0.12&0.04&0.09&0.12&1.30\\

\hline
\end{tabular}
}
\end{table*}

We compare our algorithm with the following:
improved GPCA for motion segmentation (GPCA)~\cite{gpca_motion_08},
$K$-flats (KF)~\cite{Ho03} (implemented for linear subspaces), Local
Linear Manifold Clustering (LLMC)~\cite{LLMC}, Local Subspace
Analysis (LSA)~\cite{Yan06LSA},
Multi Stage Learning (MSL)~\cite{Sugaya04},  Spectral Curvature Clustering
(SCC)~\cite{spectral_applied}, Sparse Subspace Clustering (SSC)~\cite{ssc09}, and Random Sample
Consensus
(RANSAC)~\cite{Fischler81RANSAC,Torr98geometricmotion,Tron2007}.  As before, our
algorithm is labeled LBF (Local Best-fit Flats); our algorithm with
mean shifts and using the modified choice of good neighborhood
described in section \ref{sec:eng} is labeled LBFMS.

For these algorithms, we copy the results from
http://www.vision.jhu.edu/data/hopkins155 (they are based on
experiments reported in~\cite{Tron2007} and~\cite{LLMC}) and~\cite{SCC_hopkins}, and we
just record the mean misclassification rate and the median
misclassification rate for each algorithm for any fixed $K$ (two or
three-motions) and for the different type of motions (``checker",
``traffic" and ``articulated").

\subsection{Discussion of Results}

From Table \ref{tab:error} we can see that our algorithm performs
well in various artificial instances of hybrid linear modeling (with
both linear subspace and affine subspace), and its advantage is
especially obvious with many outliers
and affine subspaces. 
The robustness to outliers is a result of our use of the $\ell_1$
error as loss function, and because of the random sampling.  Also
unlike many other methods, the proposed method natively supports
affine subspace models.

Table \ref{tab:time} shows that the running time of the proposed
algorithm is less than the running time of most other algorithms,
especially GPCA, LSA and LSCC.   The difference is large enough that
we can also use the proposed algorithm as an initialization for the
others.  The algorithm is slower than a single run of $K$-flats, but
it usually takes many restarts of $K$-flats to get a decent result.
Notice that the choice of $C$ and $p$ in our algorithm function in a
similar manner to the number of restarts in KF.

From Table \ref{tab:comp1} we can see that the local best-fit flat
algorithm works well for the data set. Of all the methods tested,
only SCC and SSC had better accuracy. However LBF ran 4 times faster
than SCC and more than 100 times faster than SSC. In many of the
cases where SSC performed better than LBF,  the $\ell_1$ energy (as
well as the $\ell_2$ energy) was lower for the labels obtained by
LBF than the labels obtained by SSC. We thus suspect that good
clustering of the Hopkins data requires additional type of
clustering (e.g., bottleneck clustering) to be combined with
subspace clustering (i.e., hybrid linear modeling).

\subsection{Initializing $K$-flats with good neighborhoods}
\label{sec:ikf} Here we demonstrate that our choice of neighborhoods
can be used to get a more robust initialization of $K$-flats.  We work
with geometric farthest insertion.  For fixed neighborhood
sizes, say of $m$ neighbors, this goes as follows:  we pick a random
point $\bx_0$ and then find the best fit flat $F_0$ for the $m$
point neighborhood of $\bx_0$.  Then we find the point $\bx_1$ in
our data farthest from $F_0$, find the best fit flat $F_1$ of the
$m$ neighborhood of $\bx_1$, and then choose the point $\bx_2$
farthest from $F_0$ and $F_1$ to continue.  We stop when we have $K$
flats; we use these as an initialization for $K$-flats.

We work on three data sets.  Data set \#1 consists of $1500$
points on three parallel $2$-planes in $\R^3$. $500$ points are
drawn from the unit square in $x,y$ plane, and then $500$ more from
the $x,y,z+.2$ plane, and then $500$ more from the $x,y,z+.4$ plane.
This data set is designed to favor the use of small neighborhoods.
The next data set is three random affine sets with 15\% Gaussian
noise and 5\% outliers, generated using the Matlab code from GPCA,
as in Section~\ref{sec:sim}. This data set is designed to favor
large neighborhood choices.  Finally, we work on a data set with
1500 points sampled from 3 planes in $\R^2$ as in
Figure~\ref{fig:nonei}. The error rates of $K$-flats with farthest
insertion initialization with fixed neighborhoods of size $10$,
$20$, $...$, $160$ are plotted against the error rate for farthest
insertion with adapted neighborhoods (searched over the same range),
averaged over 400 runs in Figure~\ref{fig:kmeansit}. Although our
method did not always beat the best fixed neighborhood, it was quite
close; and it always significantly better than the wrong fixed
neighborhood size. Both methods did significantly better than a
random initialization.

In Figure~\ref{fig:nonei} we plot the number of neighbors picked by
our algorithm for each point of a realization of data set \#3.

\subsection{Automatic determination of the number of affine sets}

In this section we show experimentally that using the elbow method
on the least squares errors of the outputs of the randomized best
fit flat method can accurately determine the number of affine
clusters.

Let $W_k$ be the total mean squared distance of a data set to the
flats returned by our algorithm with $k$ affine clusters specified;
as $k$ increases, $W_k$ decreases.  A classical method for
determining the correct number $k$ is to find the ``elbow'', or the
$k$ past which adding more clusters does not significantly decrease
the error.  We use the Second Order Difference (SOD) formulation of
this heuristic \cite{SOD}:
\begin{equation}
SOD(\ln W_k) = \ln W_{k-1} + \ln W_{k+1} - 2\ln W_k,
\label{eq:SODlnWk}
\end{equation}
Then the optimal $k$ is found by:
\begin{equation}
k_{opt} = \arg \max_k SOD(\ln W_k)  \label{eq:optlnK}.
\end{equation}

We compare SOD (LBF), i.e., SOD applying LBF, with
ALC~\cite{Ma07Compression} and GPCA ~\cite{Ma07} on a number of
artificial data sets. Similarly to Section~\ref{sec:sim}, data sets
were generated by the Matlab code borrowed from the GPCA package in
http://perception.csl.uiuc.edu/gpca with $100d$ samples from each
subspace and 0.05 Gaussian noise. For the last four experiments, we
restrict the angle between subspaces to be at least $\pi/8$ for
separation. All algorithms are given the dimension $d$ and we choose
$k_{max}=10$ in SOD (LBF).
 For ALC,
we use the oracle choice of the parameter $\epsilon$, setting it
equal the true noise level. For GPCA, we embed the data to a $d+1$
subspace by PCA and let the tolerance of rank detection be $0.05$
\cite{Vidal05, Ma07}. There is no automatic way to choose this
tolerance, so we tried different values and picked the one which
matched the ground truth the best. Each experiment is repeated 100
times and the error ($e\%$) and the average computation time $t$ (in
seconds) are recorded in Table~\ref{tab:SOD}.

\section{Conclusions and future work}
\label{sec:conclusions}
 We presented a very simple geometric method for
hybrid linear modeling based on selecting a set of local best fit
flats that minimize a global $\ell_1$ error measure.  The size of
the local neighborhoods is determined automatically using the
$\ell_2$ $\beta$ numbers; it is proven under certain geometric
conditions that good local neighborhoods exist and are found by this
method. We give extensive experimental evidence demonstrating the
state of the art accuracy and speed of the algorithm on synthetic
and real hybrid linear data.

We believe that the next step is to adapt the method for
multi-manifold clustering.  As it is, our method, while quite good
at unions of affine sets, cannot successfully handle unions of
curved manifolds.  We believe that by gluing together groups of
local best fit flats related by some smoothness conditions, we will
be able to approach the problem of clustering data which lies on
unions of smooth manifolds.

\begin{small}
\bibliographystyle{unsrt}
\bibliography{refs}
\end{small}
\end{document}